\documentclass[runningheads]{llncs}
\usepackage[T1]{fontenc}
\usepackage{graphicx,verbatim}
\usepackage{bbm}
\usepackage{amsmath}
\usepackage{amssymb}
\usepackage{graphicx}
\usepackage{arydshln}
\usepackage{hyperref} 
\usepackage{multirow}
\usepackage{amsmath}
\usepackage{amssymb}
\usepackage{booktabs}
\usepackage{breqn}
\usepackage{array}
\newcolumntype{P}[1]{>{\centering\arraybackslash}p{#1}}
\usepackage{algorithm}
\usepackage{amsfonts}
\usepackage{graphicx,wrapfig,lipsum}
\newcolumntype{P}[1]{>{\centering\arraybackslash}p{#1}}
\usepackage[table]{xcolor}
 \usepackage{appendix}
\usepackage{caption}
\usepackage{subcaption}
\usepackage{tabularx}
\usepackage{comment}
\usepackage{color}

\begin{document}
\title{Multi-Kernel Gated Decoder Adapters for Robust Multi-Task Thyroid Ultrasound under Cross-Center Shift} 

\titlerunning{Multi-Kernel Gated Decoder Adapters for Multi-Task Thyroid Ultrasound}
%
% \begin{comment} 
\author{Maziar Sabouri\inst{1,2} \and
Nourhan Bayasi\inst{2,3} \and
Arman Rahmim\inst{1,2,3,4}}
\authorrunning{M. Sabouri et al.}
\institute{Department of Astronomy and Physics, University of British Columbia, Vancouver, Canada \and
Department of Basic and Translational Research, BC Cancer Research Institute, Vancouver, Canada \and
Department of Radiology, University of British Columbia, Vancouver, Canada \and 
School of Biomedical Engineering, University of British Columbia, Vancouver, Canada 
}
% \end{comment}

% \author{Anonymized Authors} 
% \authorrunning{Anonymized Author et al.}
% \institute{Anonymized Affiliations \\
%     \email{email@anonymized.com}}
  
\maketitle             
\begin{abstract}
Thyroid ultrasound (US) automation couples two competing requirements: global, geometry-driven reasoning for nodule delineation and local, texture-driven reasoning for malignancy risk assessment. Under cross-center domain shift, these cues degrade asymmetrically, yet most multi-task pipelines rely on a single shared backbone, often inducing negative transfer. In this paper, we characterize this interference across CNN (ResNet34) and medical ViT (MedSAM) backbones, and observe a consistent trend: ViTs transfer geometric priors that benefit segmentation, whereas CNNs more reliably preserve texture cues for malignancy discrimination under strong shift and artifacts. Motivated by this failure mode, we propose a lightweight family of decoder-side adapters, the Multi-Kernel Gated Adapter (MKGA) and a residual variant (ResMKGA), which refine multi-scale skip features using complementary receptive fields and apply semantic, context-conditioned gating to suppress artifact-prone content before fusion. Across two US benchmarks, the proposed adapters improve cross-center robustness: they strengthen out-of-domain segmentation and, in the CNN setting, yield clear gains in clinical TI-RADS diagnostic accuracy compared to standard multi-task baselines.  Code and models will be released.
\end{abstract}

\section{Introduction}
Thyroid nodules are routinely evaluated with ultrasound (US) to guide follow-up, biopsy decisions, and risk stratification under the 5-point scoring system TI-RADS, with fine-needle aspiration cytology (FNAC) as the diagnostic reference standard~\cite{Tessler2017,huang2022}. While deep learning has achieved strong performance for nodule segmentation and malignancy prediction in curated datasets, reliable cross-center generalization remains a major barrier to clinical adoption. In practice, thyroid US appearance varies substantially across institutions due to scanner vendor and settings, acquisition protocol, operator technique, and frequent on-image overlays such as calipers and text markup~\cite{Pedraza_ThyroidDB_2015,Ogut2025}. These shifts often break models that appear accurate in-distribution, especially when systems are expected to support both contouring (for measurement and follow-up) and risk assessment (for triage).

A key reason is that thyroid US automation couples two qualitatively different reasoning modes. Accurate segmentation is largely geometry-driven: it requires global context to delineate uncertain boundaries, enforce plausible shape, and remain stable under speckle noise and weak edges~\cite{Zheng2023Segmentation}. In contrast, malignancy assessment is more texture-driven, relying on subtle local cues aligned with TI-RADS descriptors (e.g., echogenicity, internal echoes, micro-calcifications) that can be sensitive to scanner-dependent statistics and overlay artifacts~\cite{Kang2022SegClass}.
Critically, these cues degrade asymmetrically under domain shift: artifacts may preserve coarse shape while corrupting high-frequency texture patterns, or conversely alter apparent boundaries while leaving local texture relatively unchanged.

Despite this asymmetry, many pipelines treat segmentation and diagnosis as multi-task learning (MTL) with a single shared backbone, implicitly assuming one representation can serve both geometric and textural objectives~\cite{Lakkapragada2022NegativeTransfer}. Under cross-center shift, this assumption becomes brittle: competing supervision can drive a shared encoder toward an inductive bias that benefits one task while harming the other, resulting in negative transfer and unstable deployment behavior \cite{wang2019}. We argue that robust multi-task thyroid US systems must address these competing objectives via targeted feature refinement in the decoder and gradient-aware optimization, rather than relying on a naïve shared encoder.

In this work, we address this failure mode with two contributions. First, we empirically characterize multi-task interference under cross-center shift by evaluating both a CNN backbone (ResNet34) and a medical foundation ViT (MedSAM), revealing that geometry- and texture-driven objectives degrade differently and can conflict when forced through a shared encoder. Second, we propose a lightweight family of decoder-side adapters, the Multi-Kernel Gated Adapter (MKGA) and its residual variant ResMKGA, which improve artifact-robust skip fusion via multi-kernel refinement and context-conditioned gating (with residual channel recalibration in ResMKGA). To further reduce optimization conflict, we optionally apply gradient surgery (PCGrad) during multi-task training. Across ThyroidXL (in-domain) and DDTI (held-out cross-center external) evaluation, our approach strengthens cross-center robustness: it consistently improves out-of-domain segmentation and, in the CNN setting, yields clear gains in TI-RADS discrimination over strong shared-backbone baselines. 

\begin{figure}[t!]
    \centering
    \includegraphics[width=\linewidth]{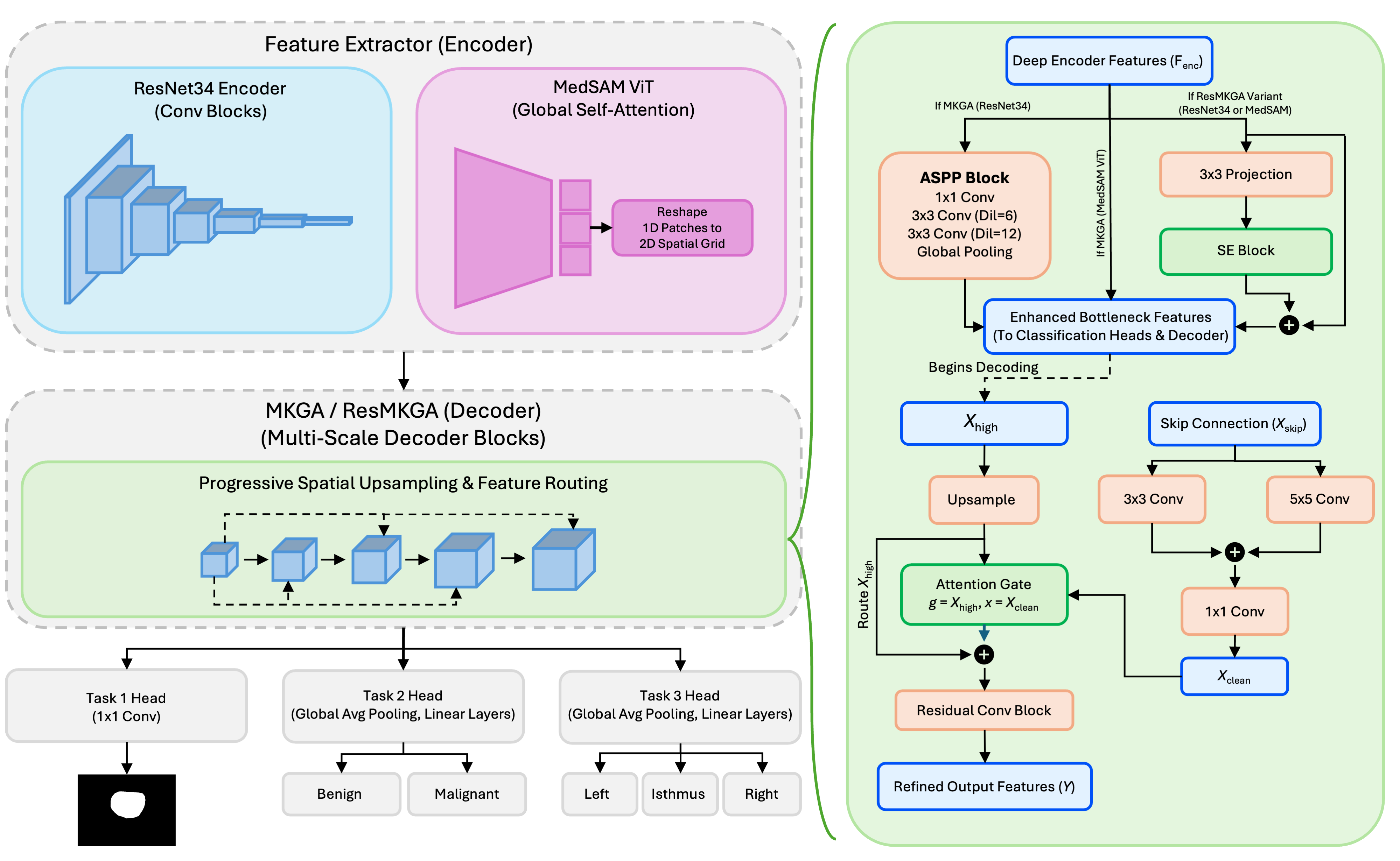}
    \caption{Overview of the proposed MKGA and ResMKGA decoders architecture.}
    \label{fig:architecture}
\end{figure}

\section{Methodology}

\noindent \textbf{2.1 Overview.}
We target robust multi-task thyroid US automation under cross-center domain shift by reducing negative transfer between geometry-driven segmentation and texture-driven classification. Our unified architecture shares a single backbone and predicts three outputs: nodule segmentation, TI-RADS malignancy, and anatomical position. To make decoding resilient to US-specific artifacts (speckle, calipers, and text overlays), we introduce a lightweight decoder adaptation family, MKGA and its residual variant ResMKGA (Fig.~\ref{fig:architecture}), which refine and selectively gate multi-scale skip features using semantic context. During training, we optimize all tasks jointly, and optionally apply gradient surgery (PCGrad) to mitigate conflicting gradients in the shared encoder.

\noindent \textbf{2.2 Backbones and Multi-Scale Decoding.}
We investigate two backbones for the unified multi-task framework. First, we adopt MedSAM~\cite{Ma2024}, a SAM-style ViT encoder~\cite{Kirillov2023SAM}, motivated by its global self-attention and strong geometric priors. Since MedSAM produces a single-scale latent map, we support multi-stage decoding by constructing a progressive pseudo-skip pyramid using learnable upsampling and projection layers to generate a sequence of aligned feature maps for the decoder stages. Second, we evaluate ResNet34~\cite{He2015} for its hierarchical receptive fields and locality bias, which are well-suited to US texture cues. In this configuration, standard encoder skip features are extracted from intermediate stages and fed into the decoder. To capture global context prior to decoding, the deepest features of the ResNet34 encoder are specifically processed through an Atrous Spatial Pyramid Pooling (ASPP) bottleneck \cite{chen2017}. We attach three task-specific heads to the unified framework: a segmentation decoder, and two global pooling heads for TI-RADS malignancy and anatomical position classification.

\noindent \textbf{2.3 Multi-Kernel Gated Adapter (MKGA).}
Direct skip fusion in a unified multi-task system can inject artifact-driven noise from shallow features into the decoder, potentially destabilizing both segmentation and classification performance. We therefore introduce MKGA, a lightweight module that refines skip features using complementary receptive fields and gates skip content conditioned on semantic context before residual fusion. Let $X_{\text{skip}}\in\mathbb{R}^{H\times W\times C_s}$ denote the encoder skip feature at a given scale, and let $X_{\text{high}}\in\mathbb{R}^{H\times W\times C_h}$ denote the upsampled deeper decoder feature aligned to the same resolution. MKGA consists of three components:

\noindent \textit{(1) Multi-kernel skip refinement.}
We apply parallel convolutions to capture multi-scale context from $X_{\text{skip}}$: a $3{\times}3$ convolution and a dilated $3{\times}3$ convolution with dilation $d{=}2$ (yielding a $5{\times}5$ receptive field). The outputs are concatenated and projected as $X_{\text{skip}}^{\text{ref}}=\phi_{1\times1}\big(\mathrm{Conv}_{3\times3}(X_{\text{skip}})\,\|\,\mathrm{Conv}_{3\times3,d=2}(X_{\text{skip}})\big)$, where $\|$ denotes channel concatenation and $\phi_{1\times1}$ is a $1{\times}1$ projection followed by normalization and ReLU.

\noindent \textit{(2) Context-conditioned gating.}
To suppress irrelevant skip activations, we use an additive attention gate conditioned on $X_{\text{high}}$. The attention map is computed as $\alpha=\sigma\!\left(\psi\!\left(\delta\!\left(W_g(X_{\text{high}})+W_s(X_{\text{skip}}^{\text{ref}})\right)\right)\right)$, where $\sigma$ is Sigmoid, $\delta$ is ReLU, and $\psi,W_g,W_s$ are $1{\times}1$ convolutions. The gated skip feature is $X_{\text{skip}}^{\text{gate}}=\alpha\odot X_{\text{skip}}^{\text{ref}}$, with $\odot$ denoting element-wise multiplication.

\noindent \textit{(3) Residual fusion.}
We fuse semantic and gated skip features via concatenation followed by a lightweight residual refinement block: $Y_{\text{MKGA}}=\mathcal{F}_{\text{res}}\big(X_{\text{high}}\,\|\,X_{\text{skip}}^{\text{gate}}\big)$, where $\mathcal{F}_{\text{res}}(\cdot)$ denotes two sequential $3{\times}3$ convolutions, each followed by normalization and ReLU.

\noindent \textbf{2.4 Residual Bottleneck Variant (ResMKGA).}
To stabilize the deepest latent representation where multi-task objectives often conflict under domain shift, we introduce ResMKGA. This module applies a residual correction to the encoder bottleneck feature $F_{\text{enc}}$ before decoding. Specifically, we compute $X_{\text{high}}^{\text{enh}}=F_{\text{enc}}+\mathrm{SE}\!\left(\phi_{3\times3}(F_{\text{enc}})\right)$, where $\phi_{3\times3}$ is a $3{\times}3$ projection and $\mathrm{SE}(\cdot)$ denotes a Squeeze-and-Excitation block \cite{hu2018} with a two-layer bottleneck MLP for channel recalibration. For ResNet34, this serves as a lightweight alternative to heavy multi-scale context heads; for MedSAM, it acts as a task-specific adapter that aligns the foundation-model features to the broader requirements of multi-task thyroid US decoding and classification.

\noindent \textbf{2.5 Training Objectives.}
We optimize the unified model with task-appropriate losses. For segmentation, we use a compound Dice and pixel-wise Cross-Entropy objective,
$\mathcal{L}_{\text{seg}}=\mathcal{L}_{\text{Dice}}+\mathcal{L}_{\text{CE}}^{\text{pixel}}$.
For malignancy and anatomical position classification, we use image-level Cross-Entropy,
$\mathcal{L}_{\text{cls}}=\mathcal{L}_{\text{CE}}^{\text{image}}$.
The overall multi-task objective is a weighted sum
$\mathcal{L}= \mathcal{L}_{\text{seg}} + \lambda_{\text{mal}}\,\mathcal{L}_{\text{mal}} + \lambda_{\text{pos}}\,\mathcal{L}_{\text{pos}}$,
with $\lambda$ set in our experiments (and PCGrad optionally applied to reduce gradient conflict).

\section{Experiments and Results}

\noindent \textbf{3.1 Datasets and Evaluation Protocol.}
We evaluate cross-center robustness using an in-domain split from the training center and an external out-of-domain benchmark from a different center. All models are trained using ThyroidXL~\cite{ThyroidXL_MICCAI2025} (11{,}635 images, 4{,}093 patients), which provides nodule masks, TI-RADS labels, and anatomical position (left/right/isthmus). We perform a patient-level split with 80\% development and 20\% in-domain test. The development set is further split into training/validation (80/20) for model selection. External generalization is assessed on DDTI~\cite{Pedraza_ThyroidDB_2015}, which is used \emph{only} for testing and contains substantial artifacts, including embedded calipers and text overlays in some cases. Following preprocessing, we split dual-view acquisitions into single-view samples and remove border artifacts, yielding 660 images. DDTI provides masks and TI-RADS labels but does not include position labels. For malignancy risk, we binarize TI-RADS into Low-Risk ($\leq 3$) vs.\ High-Risk ($\geq 4$) for both datasets \cite{Tessler2017,abdelrazik2025,kunapinun2023}.

\noindent \textbf{3.2 Implementation Details.}
Images are zero-padded to square to preserve aspect ratio and resized to $512{\times}512$; intensities are normalized using ImageNet statistics. To model inter-center variability during training, we apply appearance perturbations (Gaussian noise, Gaussian blur, multiplicative noise) and mild geometric transforms (affine, scale $<10\%$, rotation $\pm 15^\circ$, elastic deformation). Horizontal flipping is disabled to preserve laterality cues for the positioning task. Models are trained in PyTorch using AdamW (lr $=10^{-4}$, batch size $=16$) for up to 100 epochs with early stopping (patience $=15$) based on validation loss.

\noindent \textbf{3.3 Tasks, Metrics, and Statistical Testing.}
We evaluate (i) segmentation, (ii) binary TI-RADS malignancy classification, and (iii) 3-way anatomical positioning (ThyroidXL only). Segmentation uses Dice and IoU; classification uses accuracy, F1, and AUC. We test improvements over baselines using Wilcoxon signed-rank (segmentation), McNemar (accuracy), and DeLong (AUC; multi-class positioning via flattened one-hot encoding), with Benjamini--Hochberg FDR correction and significance at adjusted $p<0.05$.

\noindent \textbf{3.4 Adaptation Settings.}
For each backbone, we consider two adaptation regimes: (i) \emph{Frozen backbone}, training only the decoder and task heads to probe feature transferability; and (ii) \emph{Unfrozen adaptation}, performing full end-to-end fine-tuning for ResNet34, and parameter-efficient LoRA~\cite{Hu_LoRA_2021} for MedSAM with ranks $r\in\{4,16,32\}$ applied to attention layers while keeping the remaining transformer frozen.

\begin{figure}[t!]
    \centering
    \includegraphics[width=\linewidth]{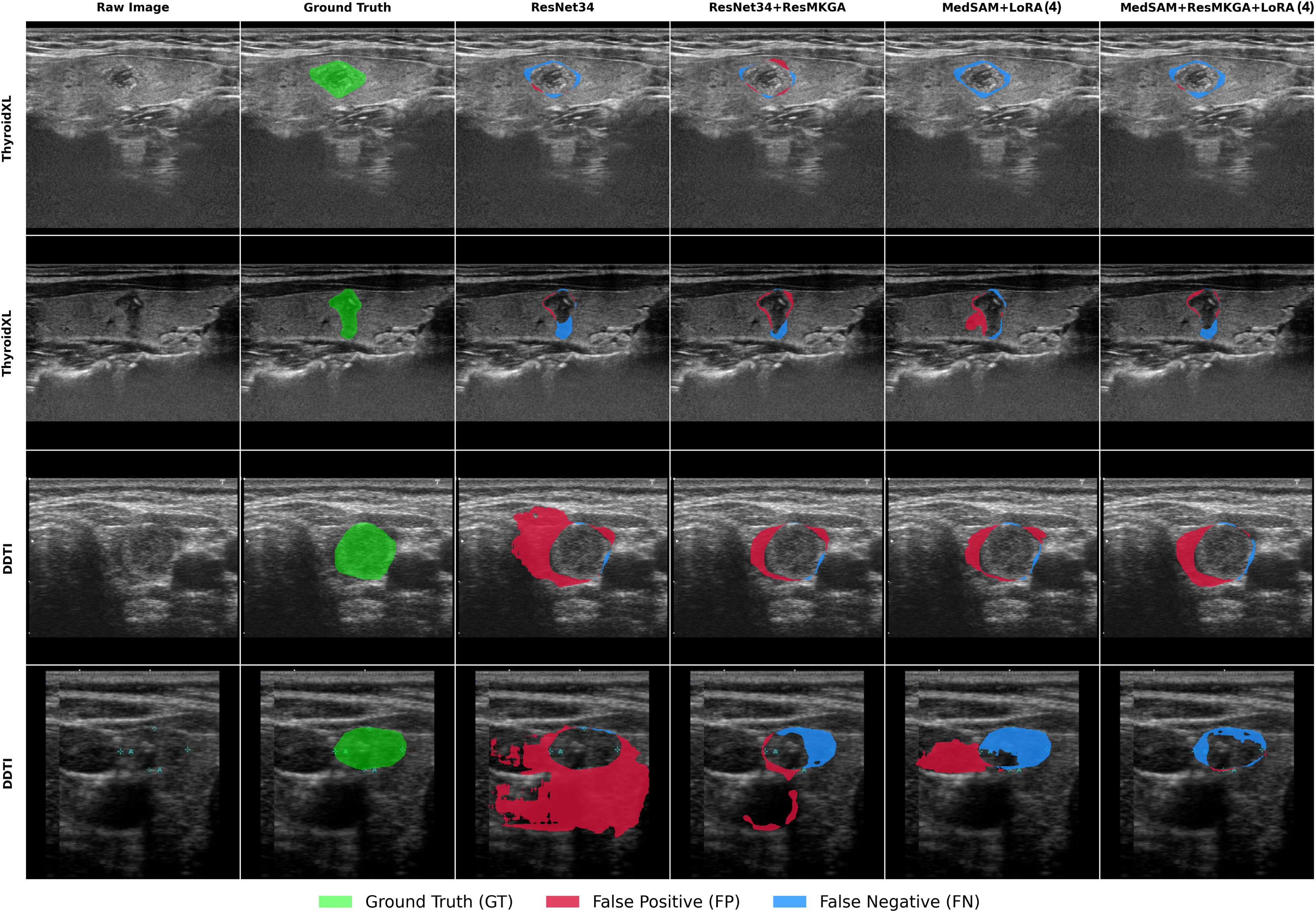}
    \caption{Qualitative segmentation comparison on ThyroidXL and DDTI datasets.}
    \label{fig:segmentation_qualitative}
\end{figure}

\noindent \textbf{3.5 Cross-Center Segmentation Robustness.}
Table~\ref{tab:main_results_standard} shows that most methods perform well in-domain on ThyroidXL (Dice $\approx 0.80$--$0.87$; best in-domain Dice: ResNet34+MKGA (Frozen), $0.869\pm0.130$), but all exhibit a clear cross-center generalization gap on DDTI. For CNNs, naïve end-to-end fine-tuning is brittle: ResNet34 (Unfrozen) drops from $0.861\pm0.130$ (in-domain) to $0.590\pm0.262$ (external). In contrast, adding MKGA/ResMKGA markedly improves external stability (DDTI Dice $0.659\pm0.236$ and $0.671\pm0.256$, respectively; $p<0.05$ vs.\ ResNet34 unfrozen). While PCGrad improves the unfrozen baseline to $0.642$ (Table~\ref{tab:pcgrad_results}), ResMKGA still surpasses this PCGrad-enhanced baseline ($p<0.05$), indicating that artifact-aware skip refinement and context-conditioned gating provide architectural robustness that reduces reliance on optimization-only fixes for segmentation. For ViTs, MedSAM+ResMKGA+LoRA ($r{=}4$) achieves the best external Dice ($0.675\pm0.226$), yet its advantage over lightweight ResNet34+ResMKGA (Unfrozen) and ResNet34+MKGA (Frozen) is not significant ($p>0.05$), suggesting the decoder adapters are the main driver of robustness across backbones. Consistent with this, increasing LoRA rank degrades external Dice ($0.675$ at $r{=}4$ to $0.655$ at $r{=}32$), supporting the hypothesis that higher-capacity adaptation overfits to source appearance rather than preserving transferable geometric structure. Qualitative examples are shown in Fig.~\ref{fig:segmentation_qualitative}.

\begin{table*}[t]
\centering
\caption{Segmentation and diagnostic (TI-RADS) performance evaluated on ThyroidXL and DDTI, and positioning on ThyroidXL only. Best in \textbf{Bold}.}
\label{tab:main_results_standard}
\resizebox{\textwidth}{!}{
\begin{tabular}{l | cc ccc ccc | cc ccc}
\toprule
 & \multicolumn{8}{c|}{\textbf{ThyroidXL (In-domain Test Set)}} & \multicolumn{5}{c}{\textbf{DDTI (External Test Set)}} \\
\cmidrule(lr){2-9} \cmidrule(lr){10-14}
\textbf{Model Architecture} & \multicolumn{2}{c}{\textbf{Segmentation}} & \multicolumn{3}{c}{\textbf{TI-RADS}} & \multicolumn{3}{c|}{\textbf{Position}} & \multicolumn{2}{c}{\textbf{Segmentation}} & \multicolumn{3}{c}{\textbf{TI-RADS}} \\
\cmidrule(lr){2-3} \cmidrule(lr){4-6} \cmidrule(lr){7-9} \cmidrule(lr){10-11} \cmidrule(lr){12-14}
 & \textbf{Dice} & \textbf{IoU} & \textbf{Acc} & \textbf{F1} & \textbf{AUC} & \textbf{Acc} & \textbf{F1} & \textbf{AUC} & \textbf{Dice} & \textbf{IoU} & \textbf{Acc} & \textbf{F1} & \textbf{AUC} \\
\midrule
\multicolumn{14}{l}{\textit{Convolutional Neural Networks (ResNet34)}} \\ 
\midrule
ResNet34 (Frozen) & 0.857 & 0.764 & 0.855 & 0.857 & 0.920 & 0.855 & 0.852 & 0.947 & 0.631 & 0.512 & 0.357 & 0.396 & 0.556 \\
ResNet34 (Unfrozen) & 0.861 & 0.772 & \textbf{0.872} & \textbf{0.873} & 0.931 & 0.864 & 0.863 & 0.953 & 0.590 & 0.466 & 0.406 & 0.466 & 0.577 \\
ResNet34 + MKGA (Frozen) & \textbf{0.869} & \textbf{0.784} & \textbf{0.873} & \textbf{0.873} & 0.915 & 0.870 & 0.868 & 0.952 & \textbf{0.671} & 0.551 & 0.430 & 0.489 & 0.600 \\
ResNet34 + MKGA (Unfrozen) & 0.862 & 0.775 & \textbf{0.872} & 0.872 & 0.922 & \textbf{0.875} & \textbf{0.870} & \textbf{0.956} & 0.659 & 0.533 & 0.632 & 0.683 & \textbf{0.642} \\
ResNet34 + ResMKGA (Frozen) & 0.853 & 0.761 & 0.862 & 0.862 & 0.923 & 0.856 & 0.855 & 0.949 & 0.596 & 0.474 & \textbf{0.676} & \textbf{0.716} & 0.609 \\
ResNet34 + ResMKGA (Unfrozen) & 0.865 & 0.777 & 0.871 & 0.872 & \textbf{0.934} & 0.865 & 0.866 & 0.951 & \textbf{0.671} & \textbf{0.553} & 0.492 & 0.557 & 0.620 \\
\midrule
\multicolumn{14}{l}{\textit{Vision Transformers (MedSAM)}} \\ 
\midrule
MedSAM (Frozen) & 0.799 & 0.690 & 0.674 & 0.679 & 0.699 & 0.727 & 0.699 & 0.832 & 0.637 & 0.507 & \textbf{0.687} & \textbf{0.709} & 0.411 \\
MedSAM + LoRA ($r=$ 4) & 0.826 & 0.727 & 0.769 & 0.767 & 0.811 & \textbf{0.831} & 0.822 & 0.928 & 0.631 & 0.504 & 0.426 & 0.497 & 0.499 \\
MedSAM + LoRA ($r=$ 16) & 0.832 & 0.732 & 0.799 & \textbf{0.800} & \textbf{0.862} & 0.824 & 0.816 & 0.925 & 0.649 & 0.518 & 0.370 & 0.426 & 0.491 \\
MedSAM + LoRA ($r=$ 32) & 0.826 & 0.726 & 0.753 & 0.752 & 0.792 & 0.818 & 0.801 & 0.926 & 0.634 & 0.510 & 0.395 & 0.461 & 0.511 \\
MedSAM + MKGA (Frozen) & 0.837 & 0.740 & 0.753 & 0.755 & 0.797 & 0.791 & 0.781 & 0.909 & 0.643 & 0.516 & 0.406 & 0.467 & 0.522 \\
MedSAM + MKGA + LoRA ($r=$ 4) & 0.850 & 0.755 & 0.799 & 0.797 & 0.851 & \textbf{0.833} & \textbf{0.829} & 0.930 & 0.651 & 0.526 & 0.430 & 0.502 & 0.455 \\
MedSAM + MKGA + LoRA ($r=$ 16) & 0.846 & 0.752 & \textbf{0.802} & \textbf{0.801} & 0.861 & 0.830 & 0.827 & \textbf{0.935} & 0.652 & 0.526 & 0.435 & 0.507 & 0.485 \\
MedSAM + MKGA + LoRA ($r=$ 32) & 0.853 & 0.760 & 0.798 & 0.788 & 0.835 & 0.827 & 0.821 & 0.929 & 0.646 & 0.519 & 0.470 & 0.543 & \textbf{0.515} \\
MedSAM + ResMKGA (Frozen) & 0.836 & 0.739 & 0.761 & 0.759 & 0.803 & 0.802 & 0.789 & 0.911 & 0.649 & 0.523 & 0.497 & 0.568 & 0.528 \\
MedSAM + ResMKGA + LoRA ($r=$ 4) & 0.853 & 0.760 & 0.800 & 0.796 & 0.851 & 0.821 & 0.813 & \textbf{0.935} & \textbf{0.675} & \textbf{0.548} & 0.463 & 0.537 & 0.478 \\
MedSAM + ResMKGA + LoRA ($r=$ 16) & \textbf{0.855} & \textbf{0.762} & 0.776 & 0.781 & 0.850 & 0.816 & 0.801 & 0.931 & 0.663 & 0.542 & 0.359 & 0.414 & 0.512 \\
MedSAM + ResMKGA + LoRA ($r=$ 32) & 0.850 & 0.758 & \textbf{0.803} & 0.795 & 0.858 & 0.830 & 0.822 & 0.934 & 0.655 & 0.531 & 0.514 & 0.585 & 0.509 \\
\bottomrule
\end{tabular}
}
\end{table*}

\begin{table*}[t!]
\centering
\caption{Impact of PCGrad on multi-task optimization. Best in \textbf{Bold}.
}
\label{tab:pcgrad_results}
\resizebox{\textwidth}{!}{
\begin{tabular}{l | cc ccc ccc | cc ccc}
\toprule
 & \multicolumn{8}{c|}{\textbf{ThyroidXL (In-domain Test Set)}} & \multicolumn{5}{c}{\textbf{DDTI (External Test Set)}} \\
\cmidrule(lr){2-9} \cmidrule(lr){10-14}
\textbf{Model Architecture} & \multicolumn{2}{c}{\textbf{Segmentation}} & \multicolumn{3}{c}{\textbf{TI-RADS}} & \multicolumn{3}{c|}{\textbf{Position}} & \multicolumn{2}{c}{\textbf{Segmentation}} & \multicolumn{3}{c}{\textbf{TI-RADS}} \\
\cmidrule(lr){2-3} \cmidrule(lr){4-6} \cmidrule(lr){7-9} \cmidrule(lr){10-11} \cmidrule(lr){12-14}
 & \textbf{Dice} & \textbf{IoU} & \textbf{Acc} & \textbf{F1} & \textbf{AUC} & \textbf{Acc} & \textbf{F1} & \textbf{AUC} & \textbf{Dice} & \textbf{IoU} & \textbf{Acc} & \textbf{F1} & \textbf{AUC} \\
\midrule
\multicolumn{14}{l}{\textit{Convolutional Neural Networks (ResNet34)}} \\ 
\midrule
ResNet34 (Unfrozen) & 0.861 & 0.772 & \textbf{0.872} & 0.873 & 0.931 & 0.864 & 0.863 & 0.953 & 0.590 & 0.466 & 0.406 & 0.466 & 0.577 \\
ResNet34 + PCGrad (Unfrozen) & 0.859 & 0.768 & \textbf{0.872} & \textbf{0.874} & \textbf{0.934} & 0.873 & \textbf{0.872} & \textbf{0.959} & 0.642 & 0.517 & 0.594 & 0.653 & 0.633 \\
\midrule
ResNet34 + MKGA (Unfrozen) & 0.862 & 0.775 & \textbf{0.872} & \textbf{0.872} & 0.922 & \textbf{0.875} & 0.870 & \textbf{0.956} & \textbf{0.659} & \textbf{0.533} & \textbf{0.632} & \textbf{0.683} & \textbf{0.642} \\
ResNet34 + MKGA + PCGrad (Unfrozen) & \textbf{0.867} & \textbf{0.779} & 0.869 & 0.870 & 0.929 & 0.873 & 0.870 & 0.954 & 0.635 & 0.510 & 0.554 & 0.616 & 0.637 \\
\midrule
ResNet34 + ResMKGA (Unfrozen) & 0.865 & 0.777 & \textbf{0.871} & \textbf{0.872} & \textbf{0.934} & 0.865 & 0.866 & 0.951 & \textbf{0.671} & \textbf{0.553} & 0.492 & 0.557 & \textbf{0.620} \\
ResNet34 + ResMKGA + PCGrad (Unfrozen) & \textbf{0.867} & \textbf{0.779} & 0.858 & 0.838 & 0.926 & \textbf{0.873} & \textbf{0.871} & \textbf{0.957} & 0.645 & 0.521 & \textbf{0.601} & \textbf{0.659} & 0.603 \\
\midrule
\multicolumn{14}{l}{\textit{Vision Transformers (MedSAM)}} \\ 
\midrule
MedSAM + LoRA ($r=$ 4) & 0.826 & 0.727 & 0.769 & 0.767 & 0.811 & \textbf{0.831} & \textbf{0.822} & 0.928 & \textbf{0.631} & \textbf{0.504} & \textbf{0.426} & \textbf{0.497} & \textbf{0.499} \\
MedSAM + LoRA ($r=$ 4) + PCGrad & \textbf{0.835} & \textbf{0.737} & \textbf{0.798} & \textbf{0.800} & \textbf{0.865} & 0.825 & 0.819 & \textbf{0.934} & 0.629 & 0.501 & 0.366 & 0.426 & 0.491 \\
\midrule
MedSAM + ResMKGA + LoRA ($r=$ 4) & 0.853 & 0.760 & \textbf{0.800} & 0.796 & 0.851 & 0.821 & 0.813 & \textbf{0.935} & \textbf{0.675} & \textbf{0.548} & 0.463 & 0.537 & \textbf{0.478} \\
MedSAM + ResMKGA + LoRA ($r=$ 4) + PCGrad & \textbf{0.858} & \textbf{0.767} & 0.799 & \textbf{0.800} & \textbf{0.858} & \textbf{0.827} & \textbf{0.825} & \textbf{0.935} & 0.658 & 0.536 & \textbf{0.512} & \textbf{0.583} & 0.437 \\
\bottomrule
\end{tabular}
}
\end{table*}

\begin{table*}[t]
\centering
\caption{Ablation studies isolating the architectural components of the proposed MKGA and ResMKGA modules. Best in \textbf{Bold}.}
\label{tab:ablation_studies}
\resizebox{\textwidth}{!}{
\begin{tabular}{l | cc ccc ccc | cc ccc}
\toprule
 & \multicolumn{8}{c|}{\textbf{ThyroidXL (In-domain Test Set)}} & \multicolumn{5}{c}{\textbf{DDTI (External Test Set)}} \\
\cmidrule(lr){2-9} \cmidrule(lr){10-14}
\textbf{Model Architecture} & \multicolumn{2}{c}{\textbf{Segmentation}} & \multicolumn{3}{c}{\textbf{TI-RADS}} & \multicolumn{3}{c|}{\textbf{Position}} & \multicolumn{2}{c}{\textbf{Segmentation}} & \multicolumn{3}{c}{\textbf{TI-RADS}} \\
\cmidrule(lr){2-3} \cmidrule(lr){4-6} \cmidrule(lr){7-9} \cmidrule(lr){10-11} \cmidrule(lr){12-14}
 & \textbf{Dice} & \textbf{IoU} & \textbf{Acc} & \textbf{F1} & \textbf{AUC} & \textbf{Acc} & \textbf{F1} & \textbf{AUC} & \textbf{Dice} & \textbf{IoU} & \textbf{Acc} & \textbf{F1} & \textbf{AUC} \\
\midrule
\multicolumn{14}{l}{\textit{Core MKGA Ablations (ResNet34 Backbone)}} \\ 
\midrule
ResNet34 + MKGA (NoGate) & 0.862 & 0.775 & 0.869 & 0.869 & 0.930 & 0.861 & 0.859 & 0.953 & \textbf{0.673} & 0.552 & 0.499 & 0.569 & 0.589 \\
ResNet34 + MKGA (NoMulti) & 0.855 & 0.768 & 0.862 & 0.863 & 0.922 & 0.874 & \textbf{0.871} & 0.955 & 0.629 & 0.509 & 0.359 & 0.402 & 0.565 \\
ResNet34 + MKGA (K1\_3) & \textbf{0.868} & \textbf{0.780} & \textbf{0.874} & \textbf{0.874} & 0.924 & 0.872 & 0.870 & 0.953 & 0.652 & 0.527 & 0.479 & 0.547 & 0.591 \\
ResNet34 + MKGA (K3\_7) & 0.862 & 0.774 & 0.869 & 0.870 & 0.922 & 0.859 & 0.855 & 0.948 & 0.604 & 0.477 & 0.497 & 0.558 & \textbf{0.661} \\
\rowcolor{gray!10} ResNet34 + MKGA (K3\_5) & 0.862 & 0.775 & 0.872 & 0.872 & 0.922 & \textbf{0.875} & 0.870 & \textbf{0.956} & 0.659 & \textbf{0.533} & \textbf{0.632} & \textbf{0.683} & 0.642 \\
\midrule
\multicolumn{14}{l}{\textit{Residual MKGA Variant Ablations (ResNet34 Backbone)}} \\ 
\midrule
ResNet34 + ResMKGA (NoSE) & 0.864 & 0.774 & \textbf{0.873} & \textbf{0.874} & \textbf{0.939} & \textbf{0.873} & \textbf{0.868} & \textbf{0.955} & 0.639 & 0.524 & \textbf{0.548} & \textbf{0.613} & 0.615 \\
\rowcolor{gray!10} ResNet34 + ResMKGA & \textbf{0.865} & \textbf{0.777} & 0.871 & 0.872 & 0.934 & 0.865 & 0.866 & 0.951 & \textbf{0.671} & \textbf{0.553} & 0.492 & 0.557 & \textbf{0.620} \\
\bottomrule
\end{tabular}
}
\end{table*}

\noindent \textbf{3.6 Cross-Center Malignancy (TI-RADS) Generalization.}
Table~\ref{tab:main_results_standard} reports TI-RADS malignancy classification. Unlike segmentation, external performance is dominated by a \emph{texture failure mode} for ViT-based encoders: MedSAM variants perform well in-domain (e.g., MedSAM+ResMKGA+LoRA $r{=}4$: Acc $0.800$, AUC $0.851$) but collapse on DDTI (AUC $\approx 0.48$--$0.50$), consistent with TI-RADS relying on subtle local, high-frequency cues that are strongly perturbed by cross-center artifacts (calipers/text) and scanner-dependent statistics. In contrast, the CNN backbone transfers substantially better under this shift. While ResNet34 (Unfrozen) achieves only 0.577 AUC on DDTI, ResNet34+MKGA reaches 0.642 AUC and yields a large, statistically significant gain in diagnostic decisions (Acc $0.406 \rightarrow 0.632$, McNemar $p<0.001$), even though the AUC increase is not significant (DeLong $p{=}0.45$). PCGrad improves the unfrozen baseline (AUC $0.577 \rightarrow 0.633$) but does not provide consistent additional benefit once MKGA/ResMKGA are introduced, indicating that decoder-side multi-kernel refinement and context-conditioned gating already suppress artifact-driven activations while preserving discriminative texture features.

\noindent \textbf{3.7 Anatomical Positioning.} 
We evaluate anatomical position only on ThyroidXL, since DDTI lacks position labels (Table~\ref{tab:main_results_standard}). CNN-based models outperform ViT variants on this task, with ResNet34+MKGA (Unfrozen) achieving the best in-domain performance (Acc $0.875$, F1 $0.870$, AUC $0.956$), consistent with CNN hierarchies better preserving global anatomical layout than patch-based attention for coarse laterality cues. Adding PCGrad yields at most a negligible change (e.g., AUC $0.957$ for ResNet34+ResMKGA+PCGrad; Table~\ref{tab:pcgrad_results}), and differences among the ResNet34 baselines and our decoder-adapter variants are not statistically significant ($p>0.05$). Importantly, this indicates that MKGA/ResMKGA improve robustness in the harder cross-center tasks (segmentation and malignancy) \emph{without} degrading positioning, supporting targeted decoder refinement as a structural solution to multi-task interference rather than relying on gradient surgery.
%Performance on the anatomical position task is evaluated exclusively on the in-domain ThyroidXL dataset, as DDTI lacks positional annotations (Table~\ref{tab:main_results_standard}). CNN-based architectures demonstrate a clear advantage over ViT models for this task, with ResNet34+MKGA (Unfrozen) achieving the highest accuracy (Acc $0.875$, F1 $0.870$, AUC $0.956$).  This is consistent with the principle that CNN feature hierarchies are inherently well-suited for capturing the global spatial and anatomical context required for position classification, whereas patch-based self-attention struggles to ground these spatial coordinates globally. While the addition of PCGrad to the ResNet34+ResMKGA framework yields a marginally higher AUC of $0.957$ (Table~\ref{tab:pcgrad_results}), hypothesis testing reveals that the performance differences between our proposed decoders, the PCGrad variants, and the standard ResNet34 (Unfrozen) baseline are not statistically significant (McNemar and DeLong $p>0.05$).  This statistical equivalence is a crucial validation of our architecture: it confirms that the MKGA and ResMKGA modules successfully solve the multi-task optimization problem without suffering from negative transfer. By preserving state-of-the-art spatial feature extraction for positioning while simultaneously driving the significant robustness gains observed in the segmentation and cross-center malignancy tasks, our decoders provide a superior structural alternative to optimization-based gradient surgery.

\noindent \textbf{3.8 Ablation Studies.} 
Table~\ref{tab:ablation_studies} isolates the architectural drivers of MKGA/ResMKGA on a ResNet34 backbone and reveals a consistent segmentation--diagnosis trade-off under cross-center shift. \textit{Gating vs.\ raw skip fusion:} Removing the context-conditioned gate (NoGate) slightly improves external segmentation (Dice $0.673$ vs.\ $0.659$, $p<0.001$) but sharply degrades TI-RADS generalization (AUC $0.589$ vs.\ $0.642$; Acc $0.499$ vs.\ $0.632$, $p<0.001$), indicating that gating is primarily a \emph{texture-preservation} mechanism that filters artifact-driven activations harmful for diagnosis, even if raw skip cues suffice for boundary delineation. \textit{Multi-kernel refinement:} Removing the multi-kernel block (NoMulti) harms both tasks (DDTI Dice $0.629$, $p=0.011$; AUC $0.565$) and causes a major collapse in diagnostic accuracy (Acc $0.359$, $p<0.001$), confirming that complementary receptive fields are necessary to capture multi-scale TI-RADS texture cues under shift. \textit{Kernel design:} Small receptive fields 1$\times$1 \& 3$\times$3 (K1\_3) retain segmentation (Dice $0.652$, $p=0.15$) but fail on diagnosis (Acc $0.479$, $p<0.001$), while overly large fields 3$\times$3 \& 7$\times$7 (K3\_7) are unstable (Dice $0.604$, $p<0.001$; Acc $0.497$, $p<0.001$) despite a marginally higher AUC (0.661, not significant). The proposed 3$\times$3 \& 5$\times$5 (K3\_5) setting provides the best balance, sustaining strong segmentation and high diagnostic robustness. \textit{Bottleneck stabilization (ResMKGA):} Ablating SE (NoSE) reduces external segmentation (Dice $0.639$ vs.\ $0.671$, $p=0.003$) with negligible AUC change (0.615 vs.\ 0.620), suggesting residual channel recalibration mainly stabilizes deep features to preserve boundary coherence under heavy shift.

\section{Conclusion}
In this paper, we study cross-center robustness for multi-task thyroid ultrasound and show that domain shift amplifies negative transfer between geometry-driven segmentation and texture-driven TI-RADS cues. We find MedSAM to transfer well for segmentation, while ResNet34 is more reliable for malignancy and positioning under artifact-heavy shift. To mitigate this within a unified model, we introduce lightweight decoder adapters (MKGA/ResMKGA) that refine and gate multi-scale skip features to suppress artifact-driven activations. These modules strengthen out-of-domain segmentation and improve cross-center TI-RADS discrimination in the CNN setting, supporting targeted decoder refinement as a practical, parameter-efficient strategy for robust clinical deployment.
%We investigated the cross-center robustness of multi-task thyroid US automation and identified a significant performance gap driven by domain shift and task interference. Our findings reveal that while ViT-based foundation models like MedSAM provide strong geometric priors for segmentation, CNNs such as ResNet34 remain more stable for extracting the high-frequency texture cues required for TI-RADS malignancy assessment and anatomical positioning, especially in the presence of artifacts like calipers and text overlays. To address these challenges within a unified framework, we proposed the Multi-Kernel Gated Adapter (MKGA) and its residual variant (ResMKGA). Our results demonstrate that these lightweight decoder modules provide essential architectural robustness by filtering out artifact-driven noise through multi-scale refinement and context-conditioned gating. Across the ThyroidXL and DDTI benchmarks, our ResMKGA-equipped models significantly outperformed standard baselines on external datasets. Overall, this work establishes that incorporating task-specific gated adapters into a unified multi-task architecture is a practical and effective design choice for achieving clinically realistic, robust thyroid US deployment.

\end{document}